\pdfoutput=1
\documentclass[11pt,a4paper]{article}
\usepackage[hyperref]{emnlp2018}

\usepackage{times}
\usepackage{url}
\usepackage{latexsym}
\usepackage{comment}
\usepackage{tikz-dependency}
\usepackage{tikz}
\usepackage{tikz-qtree}
\usepackage[british]{babel}
\usepackage{graphicx}
\usepackage{colortbl}
\usepackage{hhline}
\usepackage{url}
\usepackage{latexsym}
\usepackage{amssymb}
\usepackage{amsmath}
\usepackage{enumerate}
\usepackage{algorithm}
\usepackage{algorithmicx}
\usepackage[noend]{algpseudocode}
\usepackage{booktabs}
\usepackage{proof}
\usepackage{paralist}
\usepackage{tikz-dependency}
\usepackage{tikz} 
\usepackage{graphicx, subfigure}
\usepackage{epstopdf}

\usepackage[latin1]{inputenc}

\usepackage{enumitem}
\setitemize{noitemsep,topsep=0pt,parsep=0pt,partopsep=0pt}
\setenumerate{noitemsep,topsep=0pt,parsep=0pt,partopsep=0pt}

\newcommand{\eat}[1]{}

\makeatletter
\useshorthands{"}%
\defineshorthand{"-}{\nobreak-\bbl@allowhyphens}
\makeatother

\newcommand{\transname}[1]{\ensuremath{\mathsf{#1}}}

\newcommand{\stacktop}{{\mid}}

\mathchardef\mhyphen="2D 

\newcommand{\sh}{\transname{Shift}}
\newcommand{\re}{\transname{Reduce}}
\newcommand{\nt}{\transname{Non\mhyphen Terminal}}

\newcommand{\fin}{\transname{Finish}}

\newcommand{\squishlist}{ 
   \begin{list}{$\bullet$}
    { \setlength{\itemsep}{0pt}      \setlength{\parsep}{3pt} 
      \setlength{\topsep}{3pt}       \setlength{\partopsep}{0pt}
      \setlength{\leftmargin}{1.5em} \setlength{\labelwidth}{1em}
      \setlength{\labelsep}{0.5em} } }

\newcommand{\squishend}{
    \end{list}  }

\aclfinalcopy 
\def\aclpaperid{791} 

\title{Dynamic Oracles for Top-Down and In-Order Shift-Reduce Constituent Parsing}

\author{Daniel Fern\'{a}ndez-Gonz\'{a}lez \and Carlos G\'{o}mez-Rodr\'{i}guez\\
	Universidade da Coru\~{n}a\\
	FASTPARSE Lab, LyS Research Group, Departamento de Computaci\'{o}n \\
	Campus de Elvi\~{n}a, s/n, 15071 A Coru\~{n}a, Spain \\
  {\tt d.fgonzalez@udc.es}, {\tt carlos.gomez@udc.es}\\}

\date{}

\begin{document}
\maketitle
\begin{abstract}
We introduce novel dynamic oracles for training two of the most accurate known shift-reduce algorithms for constituent parsing: the top-down and in-order transition-based parsers. In both cases, the dynamic oracles manage to notably increase their accuracy, in comparison to that obtained by performing classic static training. In addition, by improving the performance of the state-of-the-art in-order shift-reduce parser, we achieve the best accuracy to date (92.0 F1) obtained by a fully-supervised single-model greedy shift-reduce constituent parser on the WSJ benchmark.

\end{abstract}

\section{Introduction}
The shift-reduce transition-based framework was initially introduced, and successfully adapted from the dependency formalism, into constituent parsing by \citet{sagae05}, significantly increasing 
phrase-structure parsing performance.

A shift-reduce algorithm uses a sequence of transitions to modify the content of two main data structures (a buffer and a stack) and create partial phrase-structure trees (or constituents) in the stack to finally produce a complete syntactic analysis for an input sentence, running in linear time. Initially, \citet{sagae05} suggested that those partial phrase-structure trees be built in a \textit{bottom-up} manner: two adjacent nodes already in the stack are combined under a non-terminal  to become a new constituent. This strategy was followed by many researchers \cite{Zhang2009, Zhu13, Watanabe15,Mi2015, Crabbe2015, Cross2016B, Coavoux2016, nonbinary} who managed to improve the accuracy and speed of the original Sagae and Lavie's bottom-up parser.
With this, shift-reduce algorithms have become competitive, and are the fastest alternative to perform phrase-structure parsing to date.

Some of these attempts \cite{Cross2016B,Coavoux2016,nonbinary} introduced \textit{dynamic oracles} \cite{goldberg2012dynamic}, originally designed for transition-based dependency algorithms, to bottom-up constituent parsing. They propose to use these dynamic oracles to train shift-reduce parsers instead of a traditional \textit{static oracle}. The latter follows the standard procedure that uses a gold sequence of transitions to train a model for parsing new sentences at test time. A shift-reduce parser trained with this approach tends to be  prone to suffer from error propagation (i.e. errors made in previous states are propagated to subsequent states, causing further mistakes in the transition sequence). Dynamic oracles \cite{goldberg2012dynamic} were developed to minimize the effect of error propagation by training parsers under closer conditions to those found at test time, where mistakes are inevitably made. They are designed to guide the parser through any state it might reach during learning time. This makes it possible to introduce error exploration to force the parser to  go through non-optimal states, teaching it how to recover from mistakes and lose the minimum number of gold constituents.

Alternatively, some researchers decided to 
follow a different direction
and explore non-bottom-up strategies for producing phrase-structure syntactic analysis.

On the one hand, \cite{Dyer2016,Kuncoro2017} proposed a \textit{top-down} transition-based algorithm, which creates a phrase structure tree in the stack by first choosing the non-terminal on the top of the tree, and then considering which should be its child nodes. In contrast to the bottom-up approach, this top-down strategy adds a lookahead guidance to the parsing process, while it loses rich local features from partially-built trees.

On the other hand, \citet{Liu2017} recently developed a novel strategy that finds a compromise between the strengths of top-down and bottom-up approaches, resulting in state-of-the-art accuracy. Concretely, this parser builds the tree following an \textit{in-order} traversal: instead of starting the tree from the top, it chooses the non-terminal of the resulting subtree after having the first child node in the stack. In that way 
each partial constituent tree is created in a bottom-up manner, 
but the non-terminal node is not chosen when all child nodes are in the stack (as a purely bottom-up parser does), but after the first child is considered.

\citet{Liu2017} report that the top-down approach is on par with the bottom-up strategy in terms of accuracy and the in-order parser yields the best accuracy to date on the WSJ. However, despite being two adequate alternatives to the traditional bottom-up strategy, no further work has been undertaken to improve their performance.\footnote{In parallel to this work, \citet{FriedK18} present a non-optimal dynamic oracle for training the top-down parser.}

We propose what, to our knowledge, are the first optimal dynamic oracles for both the top-down and in-order shift-reduce parsers, allowing us to train these algorithms with exploration. The resulting parsers 
outperform the existing versions trained with static oracles on the WSJ Penn Treebank \cite{marcus93} and Chinese Treebank (CTB) benchmarks \cite{Xue2005}. The version of the in-order parser trained with our dynamic oracle achieves the highest accuracy obtained so far by a single fully-supervised greedy shift-reduce system on the WSJ.

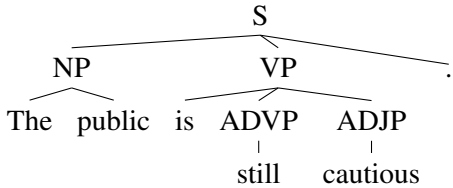
\begin{figure}[t]
\centering
\begin{tikzpicture}[level distance=0.7cm]
\tikzset{frontier/.style={distance from root=76pt}}
\Tree [.S [.{NP} [.The ] [.public ] ]
[.VP [.is ] [.ADVP [.still ] ] [.ADJP [.cautious ] ] ] [.. ] ]
\end{tikzpicture}

\caption{Simplified constituent tree, taken from English WSJ~\S{22}.}
\label{fig:ctrees}
\end{figure}

\section{Preliminaries}
The original transition system of
\citet{sagae05} parses a sentence from left to right by reading (moving) words from a \textit{buffer} to a \textit{stack}, where partial subtrees are built. This process is performed by a
sequence of $\sh$ (for reading) and $\re$ (for building) transitions that will lead the parser through different states or parser configurations until a terminal one is reached. While in the bottom-up strategy the $\re$ transition is in charge of labeling the partial subtree with a non-terminal at the same time the tree is built, \citet{Dyer2016} and \citet{Liu2017} introduce a novel transition to choose the non-terminal on top, leaving the $\re$ transition just to create the subtree under the previously decided non-terminal. We will now explain more in detail both the top-down and the in-order transition systems.

In both transition systems, parser configurations have the form {$c=\langle {\Sigma} , {i} , {f}, {\gamma}, {\alpha} \rangle$, where $\Sigma$ is a stack of constituents, $i$ is the position of the leftmost unprocessed word in the buffer (which is the next to be pushed onto the stack), $f$ is a boolean variable used by the in-order transition system to mark if a configuration is terminal or not and with no value in top-down parser configurations, $\gamma$ is the set of constituents that have already been built, and $\alpha$ is the set of non-terminal nodes that are currently in the stack.

Each constituent is represented as a tuple $(X,l,r)$, where $X$ is a non-terminal and $l$ and $r$ are integers defining its span. Constituents are composed of one or several words or constituents, and just one non-terminal node on top. Each word $w_i$ is represented as $(w, i, i+1)$. To define our oracles, we will need to represent each non-terminal node of the tree as $(X,j)$, where $j$ has the value of $i$ when $X$ is included in the stack and is used to keep them in order.\footnote{When two or more non-terminals share their labels within the tree, we use a secondary index to make them unique.}

For instance, the phrase-structure tree in Figure~\ref{fig:ctrees} can be decomposed as the following set of gold constituents: \{(S, 0, 6), (NP, 0, 2), (VP, 2, 5), (ADVP, 3, 4), (ADJP, 4, 5)\}. In addition, the ordered set of gold non-terminal nodes added to the stack while following a top-down strategy will be \{(S, 0), (NP, 0), (VP, 2), (ADVP, 3), (ADJP, 4)\} and, according to an in-order approach, \{(NP, 1), (S, 2), (VP, 3), (ADVP, 4), (ADJP, 5)\}. It is worth mentioning that the index of non-terminal nodes in the top-down method is the same as the leftmost span index of the constituent that it will produce. However, this does not hold in the in-order approach, as the leftmost child is fully processed before the node is added to the stack, so the index for the node will point to the leftmost span index of the second leftmost child.

Note that the information about the span of a constituent, the set of predicted constituents $\gamma$ and the set $\alpha$ of predicted non-terminal nodes in the stack is not used by the original top-down and in-order parsers. However, we need to include it in parser configurations at learning time to allow an efficient implementation of the proposed dynamic oracles.

Given an input string $w_0 \cdots w_{n-1}$, the in-order parsing process starts at the initial configuration  $c_s(w_0 \ldots w_{n-1}) = \langle [\ ], 0 , \mathit{false}, \{ \}, \{ \} \rangle$ and, after applying a sequence of transitions, it ends in a terminal configuration $\langle (S,0,n) , n , \mathit{true}, \gamma, \alpha \rangle$, where $n$ is the number of words in the input sentence. The top-down transition system shares the same form for the initial and terminal configurations, except for the fact that variable $f$ has no value in both cases.

\begin{figure*}

\small
\begin{tabbing}
\hspace{0.5cm}\=\hspace{2.5cm}\= \kill
\> \sh: 
\> \ \ \ \ \ \ \ $\langle {\Sigma}, {i}, /, \gamma, \alpha  \rangle
\Rightarrow \langle {\Sigma \stacktop (w_i, i, i+1)} , {i+1}, /, \gamma \cup \{ (w_i, i, i+1) \}, \alpha \rangle$\\[2mm]
\> \nt-X:
\>  \ \ \ \ \ \ \ $\langle {\Sigma}, i , /, \gamma, \alpha \rangle
\Rightarrow \langle {\Sigma \stacktop (X, i) }, i, /, \gamma, \alpha \cup \{ (X, i) \}  \rangle$ { }\\[2mm]
\> \re:
\>   \ \ \ \ \ \ \   $\langle {\Sigma \stacktop (X, j) \stacktop (Y_{1},m_0, m_1) \stacktop ... \stacktop (Y_k, m_{k-1}, m_k)}, i , /, \gamma, \alpha \rangle$\\
\>  \ \ \ \ \ \ \  \hspace{6cm}  $\Rightarrow \langle {\Sigma \stacktop (X, j, m_k), i, /, \gamma \cup \{ (X, j, m_k)\}, \alpha \setminus \{ (X, j) \} } \rangle$ { } 
\end{tabbing}

\caption{Transitions of a top-down constituent parser.}
\label{fig:transitions}
\end{figure*}

Figure~\ref{fig:transitions}
shows
the available transitions in the top-down algorithm.
In particular, the $\sh$ transition  
moves the first (leftmost) word in the buffer to the stack;
the $\nt$-X transition pushes onto the stack the non-terminal node X that should be on top of a coming constituent, and the $\re$ transition
pops the topmost stack nodes until the first non-terminal node appears (which is also popped) and combines them into a constituent with this non-terminal node as their parent,
pushing this new constituent into the stack.
Note that every reduction action will add a new constituent to $\gamma$ and remove a non-terminal node from $\alpha$, and every $\nt$ transition will include a new non-terminal node in $\alpha$.
Figure~\ref{fig:trans} shows the top-down transition sequence that produces
the phrase-structure tree in Figure~\ref{fig:ctrees}.

\begin{figure}
\begin{center}
\small

\begin{tabular}{@{\hskip 0pt}l@{\hskip 0pt}c@{\hskip 0.5pt}r@{\hskip 0pt}}
\hline\noalign{\smallskip}
\scriptsize{Transition} & $\Sigma$ & \scriptsize{Buffer}  \\
\noalign{\smallskip}\hline\noalign{\smallskip}
 & [ ] & [ The, ...] \\
\textsc{NT}$_{\textsc{S}}$ & [ S ] & [ The, ...] \\
\textsc{NT}$_{\textsc{NP}}$ & [ S, NP ] & [ The, ...] \\
\textsc{SH} & [ S, NP, The ] & [ public, ...] \\
\textsc{SH} & [ S, NP, The, public ] & [ is, ...] \\
\textsc{RE} & [ S, \textbf{NP} ] & [ is, ...] \\
\textsc{NT}$_{\textsc{VP}}$ & [ S, \textbf{NP}, VP ] & [ is, ...] \\
\textsc{SH} & [ S, \textbf{NP}, VP, is ] & [ still, ...] \\
\textsc{NT}$_{\textsc{ADVP}}$ & [ S, \textbf{NP}, VP, is, ADVP ] & [ still, ...] \\
\textsc{SH} & [ S, \textbf{NP}, VP, is, ADVP, still ] & [ cautious, ...] \\
\textsc{RE} & [ S, \textbf{NP}, VP, is, \textbf{ADVP} ] & [ cautious, ...] \\
\textsc{NT}$_{\textsc{ADJP}}$ & [ S, \textbf{NP}, VP, is, \textbf{ADVP}, ADJP ] & [ cautious, ...] \\
\textsc{SH} & [S, \textbf{NP},VP, is, \textbf{ADVP}, ADJP, cautious] & [ \textbf{.} ] \\
\textsc{RE} & [ S, \textbf{NP}, VP, is, \textbf{ADVP}, \textbf{ADJP} ] & [ \textbf{.} ] \\
\textsc{RE} & [ S, \textbf{NP}, \textbf{VP} ] & [ \textbf{.} ] \\
\textsc{SH} & [ S, \textbf{NP}, \textbf{VP}, \textbf{.} ] & [ ] \\
\textsc{RE} & [ \textbf{S} ] & [ ] \\
\noalign{\smallskip}\hline
\end{tabular}

\caption{Transition sequence for producing the constituent tree in Figure~\ref{fig:ctrees} using
a top-down parser. SH = $\sh$, NT$_{\textsc{X}}$ = $\nt$-X and RE = $\re$. Already-built constituents are marked in bold.} \label{fig:trans}       
\vspace*{13pt}
\end{center}
\end{figure}

\begin{figure*}

\small
\begin{tabbing}
\hspace{0cm}\=\hspace{2.5cm}\= \kill
\> \sh: 
\> \ \ \ \ \ \ \ $\langle {\Sigma}, {i}, \mathit{false}, \gamma, \alpha  \rangle
\Rightarrow \langle {\Sigma \stacktop (w_i, i, i+1)} , {i+1}, \mathit{false}, \gamma \cup \{ (w_i, i, i+1) \}, \alpha \rangle$\\[2mm]
\> \nt-X:
\>  \ \ \ \ \ \ \ $\langle {\Sigma}, i , \mathit{false}, \gamma, \alpha \rangle
\Rightarrow \langle {\Sigma \stacktop (X, i) }, i, \mathit{false}, \gamma, \alpha \cup \{ (X, i) \}  \rangle$ { }\\[2mm]
\> \re:
\>   \ \ \ \ \ \ \   $\langle {\Sigma \stacktop (Y_{1},m_0, m_1) \stacktop (X, j) \stacktop ... \stacktop (Y_k, m_{k-1}, m_k)}, i , \mathit{false}, \gamma, \alpha \rangle$\\
\>  \ \ \ \ \ \ \  \hspace{5cm}  $\Rightarrow \langle {\Sigma \stacktop (X, m_0, m_k), i, \mathit{false}, \gamma \cup \{ (X, m_0, m_k)\}, \alpha \setminus \{ (X, j) \} } \rangle$ { }\\[2mm]
\> \fin:
\>   \ \ \ \ \ \ \   $\langle {(S,0,n)}, n, \mathit{false}, \gamma, \alpha \rangle
\Rightarrow \langle {(S,0,n)}, n, \mathit{true}, \gamma, \alpha \rangle$ { }
\end{tabbing}

\caption{Transitions of a in-order constituent parser.}
\label{fig:transitions2}
\end{figure*}

In Figure~\ref{fig:transitions2}
we describe the available transitions in the in-order algorithm. The $\sh$, $\nt$-X and $\re$ transitions have the same behavior as defined for the top-down transition system, except that the $\re$ transition not only pops stack nodes until finding a non-terminal node (also removed from the stack), but also the node below this non-terminal node, and combines them into a constituent spanning all the popped nodes with the non-terminal node on top. And, finally, a $\fin$ transition is also available to end the parsing process. Figure~\ref{fig:trans2} shows the in-order transition sequence 
that outputs the constituent tree in Figure~\ref{fig:ctrees}.

\begin{figure}
\begin{center}
\small

\begin{tabular}{@{\hskip 0pt}l@{\hskip 0pt}c@{\hskip 0.2pt}r@{\hskip 0pt}}
\hline\noalign{\smallskip}
\scriptsize{Transition} & $\Sigma$ & \scriptsize{Buffer}  \\
\noalign{\smallskip}\hline\noalign{\smallskip}
 & [ ] & [ The, ...] \\
\textsc{SH} & [ The ] & [ public, ...] \\
\textsc{NT}$_{\textsc{NP}}$ & [ The, NP ] & [ public, ...] \\
\textsc{SH} & [ The, NP, public ] & [ is, ...] \\
\textsc{RE} & [ \textbf{NP} ] & [ is, ...] \\
\textsc{NT}$_{\textsc{S}}$ & [ \textbf{NP}, S ] & [ is, ...] \\
\textsc{SH} & [ \textbf{NP}, S, is ] & [ still, ...] \\
\textsc{NT}$_{\textsc{VP}}$ & [ \textbf{NP}, S, is, VP ] & [ still, ...] \\
\textsc{SH} & [ \textbf{NP}, S, is, VP, still ] & [ cautious, ...] \\
\textsc{NT}$_{\textsc{ADVP}}$ & [ \textbf{NP}, S, is, VP, still, ADVP ] & [ cautious, ...] \\
\textsc{RE} & [ \textbf{NP}, S, is, VP, \textbf{ADVP} ] & [ cautious, ...] \\
\textsc{SH} & [ \textbf{NP}, S, is, VP, \textbf{ADVP}, cautious ] & [ \textbf{.} ] \\
\textsc{NT}$_{\textsc{ADJP}}$ & [\textbf{NP}, S, is,VP, \textbf{ADVP}, cautious, ADJP] & [ \textbf{.} ] \\
\textsc{RE} & [ \textbf{NP}, S, is, VP, \textbf{ADVP}, \textbf{ADJP} ] & [ \textbf{.} ] \\
\textsc{RE} & [ \textbf{NP}, S, \textbf{VP} ] & [ \textbf{.} ] \\
\textsc{SH} & [ \textbf{NP}, S, \textbf{VP}, \textbf{.} ] & [ ] \\
\textsc{RE} & [ \textbf{S} ] & [ ] \\
\textsc{FI} & [ \textbf{S} ] & [ ] \\
\noalign{\smallskip}\hline
\end{tabular}

\caption{Transition sequence for producing the constituent tree in Figure~\ref{fig:ctrees} using
an in-order parser. SH = $\sh$, NT$_{\textsc{X}}$ = $\nt$-X, RE = $\re$ and FI = $\fin$. Already-built constituents are marked in bold.} \label{fig:trans2}    
\vspace*{13pt}
\end{center}
\end{figure}

The standard procedure to train a greedy shift-reduce parser  consists of training a classifier to approximate an \textit{oracle}, which chooses optimal transitions with respect to gold parse trees. This classifier will greedily choose which transition sequence the parser should apply at test time.

Depending on the strategy used for training the parser, oracles can be static or dynamic. A static oracle trains the parser only on gold transition sequences, while a dynamic one can guide the parser through any possible transition path, allowing the exploration of non-optimal sequences. 

\section{Dynamic Oracles}
Previous work such as \cite{Cross2016B,Coavoux2016,nonbinary} has introduced and successfully applied dynamic oracles for bottom-up phrase-structure parsing. We present dynamic oracles for training the top-down and in-order transition-based constituent parsers.

\citet{goldberg2012dynamic} show that implementing a dynamic oracle reduces to defining a \textit{loss function} 
on configurations to measure the distance from the best tree they can produce to the gold parse. This allows us to compute 
which transitions will lead the parser to configurations where the minimum number of mistakes are made.

\subsection{Loss function}
According to \citet{nonbinary}, we can define a loss function in constituent parsing as follows: given a parser configuration $c$ and a gold tree $t_G$, a loss function $\ell(c)$ is implemented as the minimum Hamming loss between $t$ and $t_G$, ($\mathcal{L}(t,t_G)$), where $t$ is the already-built tree of a configuration $c'$ reachable from $c$ (written as $c \rightsquigarrow t$). This Hamming loss is computed as the size of the symmetric difference between the set of constituents $\gamma$ and $\gamma_G$ in the trees $t$ and $t_G$, respectively. Therefore, the loss function is defined as:
\[ \ell(c) = \min_{\gamma | c \rightsquigarrow \gamma} \mathcal{L}(\gamma,\gamma_G) = |\gamma_G \setminus \gamma| + |\gamma \setminus \gamma_G| \] 
\noindent and, according to the authors, it can be efficiently computed for a non-binary bottom-up transition system by counting the individually unreachable arcs from configuration $c$ ($|{\mathcal{U}(c,{\gamma}_G)}|$) plus the erroneous constituents created so far ($|\gamma_c \setminus \gamma_G|$):
\[ \ell(c) = \min_{\gamma | c \rightsquigarrow \gamma} \mathcal{L}(\gamma,\gamma_G) = |{\mathcal{U}(c,{\gamma}_G)}| + |\gamma_c \setminus \gamma_G| \] 
\noindent We adapt the latter to efficiently implement a loss function for the top-down and in-order strategies. 

While in bottom-up parsing constituents are created at once by a $\re$ transition, in the other two approaches a $\nt$ transition begins the process by naming the future constituent and a $\re$ transition builds it by setting its span and children. Therefore, a $\nt$ transition that deviates from the non-terminals expected in the gold tree will eventually produce a wrong constituent in future configurations, so it should be penalized. Additionally, a sequence of gold $\nt$ transitions may also lead to a wrong final parse if they are applied in an incorrect order. Then, the computation of the Hamming loss in top-down and in-order phrase-structure parsing adds two more terms to the bottom-up loss expression: (1) the number of predicted non-terminal nodes that are currently in the stack ($\alpha_c$),\footnote{Note that we only consider predicted non-terminal nodes still in the stack, since wrong non-terminal nodes that have been already reduced are included in the loss as erroneous constituents.} but not included in the set of gold non-terminal nodes ($\alpha_G$), and (2) the number of gold non-terminal nodes in the stack that are out of order with respect to the order needed in the gold tree:
\[ \ell(c) = \min_{\gamma | c \rightsquigarrow \gamma} \mathcal{L}(\gamma,\gamma_G) = |{\mathcal{U}(c,{\gamma}_G)}| + |\gamma_c \setminus \gamma_G|\]\[ + |\alpha_c \setminus \alpha_G| + \mathit{out\_of\_order}(\alpha_c , \alpha_G) \]
This loss function is used to implement a dynamic oracle that, when given any parser configuration, will return the set of transitions $\tau$ that do not increase the overall loss (i.e., $\ell( \tau(c) ) - \ell(c) = 0 $), leading the system through optimal configurations 
that minimize
Hamming loss with respect to $t_G$.

As suggested by \cite{Coavoux2016,nonbinary}, \textit{constituent reachability} can be used to efficiently compute the first term of the symmetric difference ($|\gamma_G \setminus \gamma|$), by simply counting the gold constituents that are individually unreachable from configuration $c$, as we describe in the next subsection. 

The second and third terms of the loss ($|\gamma_c \setminus \gamma_G|$ and $|\alpha_c \setminus \alpha_G|$) can be trivially computed and are used to penalize false positives (extra erroneous constituents) so that final F-score is not harmed due to the decrease of precision, as pointed out by \cite{Coavoux2016,nonbinary}. Note that it is crucial that the creation of non-gold $\nt$ transitions is avoided, since these might not affect the creation of gold constituents, however, they will certainly lead the parser to the creation of extra erroneous constituents in future steps. 

Finally, the function $\mathit{out\_of\_order}$ of the last term can be implemented by computing the \textit{longest increasing subsequence} of gold non-terminal nodes in the stack, where the order relation is given by the order of non-terminals (provided by their associated index) in the transition sequence that builds the gold tree (this order is unique, as none of our two parsers of interest have spurious ambiguity). Obtaining the longest increasing subsequence is a well-known problem solvable in time $O$(n $log$ n) \cite{Fredman1975}, where $n$ denotes the length of the input sequence. Once we have the largest possible subsequence of gold non-terminal nodes in our configuration's stack that is compatible with the gold order, the remaining ones give us the number of erroneous constituents that we will unavoidably generate, even in the best case, due to building them in an incorrect order.

We will prove below that this loss formulation returns the exact loss and the resulting dynamic oracle is correct.

\subsection{Constituent reachability}

We now show how the computation of the set of reachable constituents developed for bottom-up parsing in \cite{Coavoux2016,nonbinary}  can be extended to deal with the top-down and in-order strategies.

\paragraph{Top-down transition system}
Let ${\gamma_G}$  and ${\alpha_G}$ be the set of gold constituents and the set of gold non-terminal nodes, respectively, for our current input. We say that a gold constituent $(X, l, r) \in {\gamma_G}$ is reachable from a configuration $c=\langle {\Sigma} , {j} , \mathit{/}, {\gamma_c}, {\alpha_c}\rangle$ with $\Sigma = [(Y_p, i_p, i_{p-1})\cdots(Y_{2}, i_{2}, i_{1})|(Y_1, i_{1}, j)]$, and it is included in the set of \textit{individually reachable constituents} $\mathcal{R}(c,{\gamma}_G)$, iff it satisfies one of the following conditions:}\footnote{Please note that elements from the stack can be an already-built constituent, a shifted word or a non-terminal node. Therefore, $(Y_p, i_p, i_{p-1})$, $(Y_{2}, i_{2}, i_{1})$ and $(Y_1, i_1, j)$ should be represented as $(Y_p, i_{p-1})$, $(Y_{2}, i_{1})$ and $(Y_1, j)$, respectively, when they are non-terminal nodes. We omit this for simplicity.}
\begin{enumerate}
\item $(X, l, r) \in \gamma_c$ (i.e. it has already been created and, therefore, it is reachable by definition).
\item $j \leq l < r \wedge (X, l) \notin \alpha_c$ (i.e. the words dominated by the gold constituent are still in the buffer and the non-terminal node that begins its creation has not been added to the stack yet; therefore, it can be still created after pushing the correct non-terminal node and shifting the necessary words). 
\item $l \in \{i_k \mid 1 \le k \le p\} \wedge j \leq r \wedge {(X, l)} \in \alpha_c $
(i.e. its span is partially or completely in the stack and the corresponding non-terminal node was already added to the stack, then, by shifting more words or/and reducing, the constituent can still be created).
\end{enumerate}

\paragraph{In-order transition system}
Let ${\gamma_G}$  and ${\alpha_G}$ be the set of gold constituents and the set of gold non-terminal nodes, respectively, for our current input. We say that a gold constituent $(X, l, r) \in {\gamma_G}$ is reachable from a configuration $c=\langle {\Sigma} , {j} , \mathit{false}, {\gamma_c}, {\alpha_c}\rangle$ with $\Sigma = [(Y_p, i_p, i_{p-1})\cdots(Y_{2}, i_{2}, i_{1})|(Y_1, i_{1}, j)]$, and it is included in the set of \textit{individually reachable constituents} $\mathcal{R}(c,{\gamma}_G)$, iff it satisfies one of the following conditions:
\begin{enumerate}
\item $(X, l, r) \in \gamma_c$ (i.e. it has already been created).
\item $j \leq l < r$ (i.e. the constituent is entirely in the buffer, then it can be still built).

\item $l \in \{i_k \mid 1 \le k \le p\} \wedge j \leq r \wedge {(X, m)} \notin \alpha_c $
(i.e. its first child is still a 
totally- or
partially-built constituent on top of the stack 
and the non-terminal node has not been created yet; therefore, it has to wait till the first child is completed 
(if it is still pending)
and, then, it can be still created by pushing onto the stack the correct non-terminal node and shifting more words if necessary).

\item $l \in \{i_k \mid 1 \le k \le p\} \wedge j \leq r \wedge {(X, m)} \in \alpha_c \wedge \exists (Y,l,m) \in \Sigma $
(i.e. its span is partially or completely in the stack, and its first child (which is an alredy-built constituent) and the non-terminal node assigned are adjacent, thus, by shifting more words or/and reducing, the constituent can still be built).
\end{enumerate}

\noindent In both transition systems, the set of \textit{individually unreachable constituents} $\mathcal{U}(c,{\gamma}_G)$ with respect to the set of gold constituents ${\gamma}_G$ can be easily computed as $ {\gamma}_G \setminus \mathcal{R}(c,{\gamma}_G)$ and will contain the gold constituents that can no longer be built.

\subsection{Correctness} 

We will now prove that the above expression of $\ell(c)$ indeed provides the minimum possible Hamming loss to the gold tree among all the trees that are reachable from configuration $c$. This implies correctness (or optimality) of our oracle.

To do so, we first show that both algorithms are constituent-decomposable. This 
amounts to saying
that if we take a set of $m$ constituents that are tree-compatible (can appear together in a constituent tree, meaning that no pair of constituent spans overlap unless one is a subset of the other) 
and individually reachable from a configuration $c$, then the set is also reachable as a whole.

We prove this by induction on $m$. The base case ($m=1$) is trivial. Let us suppose that constituent-decomposability holds for any set of $m$ tree-compatible constituents. We will show that it also holds for any set $T$ of $m+1$ tree-compatible constituents. 

Let $(X,l,r)$ be one of the constituents in $T$ such that $r = \min \{ r' \mid (X',l',r') \in T \}$ and $l = \max \{ l' \mid (X',l',r) \in T\}$. Let $T' = T \setminus \{(X,l,r)\}$. Since $T'$ has $m$ constituents, by induction hypothesis, $T'$ is a reachable set from configuration $c$.

Since $(X,l,r)$ is individually reachable by hypothesis, it must satisfy at least one of the conditions for constituent reachability. As these conditions are different for each particular algorithm, we continue the proof separately for each:

\paragraph{Top-down constituent-decomposability}

In this case, we enumerated three constituent reachability conditions, so we divide the proof into three cases: 

If the first condition holds, then the constituent $(X,l,r)$ has already been created in $c$. Thus, it will still be present after applying any of the possible transition sequences that build $T'$ starting from $c$. Hence, $T = T' \cup \{(X,l,r)\}$ is reachable from $c$.

If the second condition holds, then $j \le l < r$ and the constituent $(X,l,r)$ can be created by $l-j$ $\sh$ transitions, followed by one $\nt$ transition, $r-l$ $\sh$ transitions and one $\re$ transition. This will leave the parser in a configuration whose value of $j$ is $r$, and where stack elements with left span index $\le l$ (apart from those referencing the new non-terminal and its leftmost child) have not changed. Thus, constituents of $T'$ are still individually reachable in this configuration, as their left span index is either $\ge r$ (and then they meet the second reachability condition) or $\le l$ (and then they meet the third), so $T$ is reachable from $c$.

Finally, if the third condition holds, then we can create $(X,l,r)$ by applying $r-j$ $\sh$ transitions followed by a sequence of $\re$ transitions stopping when we obtain $(X,l,r)$ on the stack (this will always happen after a finite number of such transitions, as the reachability condition guarantees that $l$ is the left span index of some constituent already on the stack, and that $(X,l)$ is on the stack). Following the same reasoning as in the previous case regarding the resulting parser configuration, we conclude that $T$ is reachable from $c$. 

With this we have shown the induction step, and thus constituent decomposability for the top-down parser.

\paragraph{In-order constituent decomposability}

The in-order parser has four constituent reachability conditions. Analogously to the previous case, we prove the reachability of $T$ by case analysis.

If the first condition holds, then we have a situation where the constituent $(X,l,r)$ has already been created in $c$, so reachability of $T$ follows from the same reasoning as for the first condition in the top-down case.

If the second condition holds, we have $j \le l < r$ and the constituent $(X,l,r)$ can be created by $l-j+1$ $\sh$ transitions (where the last one shifts a word that will be assigned as left child of the new constituent), followed by the relevant $\nt$-X transition, $r-l-1$ more $\sh$ transitions and one $\re$ transition. After this, the parser will be in a configuration where $j$ takes the value $r$, where we can use the same reasoning as in the second condition of the top-down parser to show that all constituents of $T'$ are still reachable, proving reachability of $T$.

For the third condition, the proof is analogous but the combination of transitions that creates the non-terminal starts with a sequence composed of $\re$ transitions (when there is a non-terminal at the top of the stack) or $\nt$-Y transitions for arbitrary $Y$ (when the top of the stack is a constituent) until the top node on the stack is a constituent with left span index $l$
(this ensures that the constituent at the top of the stack can serve as leftmost child for our desired constituent), followed by a $\nt$-X, 
$r-j$
$\sh$ transitions and one $\re$ transition.

Finally, for the fourth condition, the reasoning is again analogous, but the computation leading to the non-terminal starts with as many $\re$ transitions as non-terminal nodes located above $(X,m)$ in the stack (if any). If we call $j$ the index associated to the resulting transition, then it only remains to apply $r-j$ $\sh$ transitions followed by a $\re$ transition.

\paragraph{Optimality}

With this, we have shown constituent decomposability for both parsing algorithms. This means that, for a configuration $c$, and a set of constituents that are individually reachable from $c$, there is always some computation that can build them all. This facilitates the proof that the loss function is correct. 

To finish the proof, we observe the following:

\begin{itemize}
\item Let $c'$ be a final configuration reachable from $c$. The set $(\gamma_{c'} \setminus \gamma_G)$, representing erroneous constituents that have been built, will always contain at least $|\gamma_c \setminus \gamma_G|$, as the algorithm never deletes constituents.
\item In addition, $c'$ will contain one erroneous constituent for each element of $(\alpha_c \setminus \alpha_G)$, as once a non-terminal node is on the stack, there is no way to reach a final configuration without using it to create an erroneous constituent. 
Note that these erroneous constituents do not overlap those arising from the previous item, as $\gamma_c$ stores already-built constituents and $\alpha_c$ non-terminals that have still not been used to  build a constituent.
\item Given a subset $\mathcal{S}$ of $\mathcal{R}(c,{\gamma}_G)$, the previously shown constituent decomposability property implies that there exists at least one transition sequence starting from $c$ that generates the tree $\mathcal{S} \cup (\gamma_c \setminus \gamma_G) \cup E$, where $E$ is a set of erroneous constituents containing one such constituent per element of $(\alpha_c \setminus \alpha_G)$. This tree has loss $|t_G|-(|\gamma_c \cup \mathcal{S}|) + |\gamma_c \setminus \gamma_G| + |\alpha_c \setminus \alpha_G|$. The term $|t_G|-(|\gamma_c \cup \mathcal{S}|)$ corresponds to missed constituents (gold constituents that have not been already created and are not created as part of $\mathcal{S}$), the other two to erroneous constituents.
\item As we have shown that the erroneous constituents arising from $(\gamma_{c'} \setminus \gamma_G)$ and $(\alpha_c \setminus \alpha_G)$ are unavoidable, computations yielding a tree with minimum loss are those that maximize $|\gamma_c \cup \mathcal{S}|$ in the previous term. In general, the largest possible $|\mathcal{S}|$ is for $\mathcal{S} = \mathcal{R}(c,{\gamma}_G)$. In that case, we would correctly generate every reachable constituent and the loss would be 
\[ \ell(c) = |{\mathcal{U}(c,{\gamma}_G)}| + |\gamma_c \setminus \gamma_G|\]\[ + |\alpha_c \setminus \alpha_G| \]
However, we additionally want to generate constituents in the correct order, and this may not be possible if we have already shifted some of them into the stack in a wrong order. The function $\mathit{out\_of\_order}$ gives us the number of reachable constituents that are lost for this cause in the best case. Thus, indeed, the expression
\[ \ell(c) = |{\mathcal{U}(c,{\gamma}_G)}| + |\gamma_c \setminus \gamma_G|\]\[ + |\alpha_c \setminus \alpha_G| + \mathit{out\_of\_order}(\alpha_c , \alpha_G) \]
provides the minimum loss from configuration $c$.
\end{itemize}

\section{Experiments}
\subsection{Data}
We test the two proposed approaches on 
two widely-used benchmarks
for constituent parsers: the Wall Street Journal (WSJ) sections of
the English Penn Treebank\footnote{Sections  2-21 are used as training data, Section 22 for development and Section 23 for testing} \cite{marcus93} and version 5.1 of the Penn Chinese Treebank (CTB)\footnote{Articles 001- 270 and 440-1151 are taken for training, articles 301-325 for system development, and articles 271-300 for final testing} \cite{Xue2005}. We use the same predicted POS tags and pre-trained word embeddings as \citet{Dyer2016} and \citet{Liu2017}.

\subsection{Neural Model}
To perform a fair comparison, we define the novel dynamic oracles on the original implementations of the top-down parser by \newcite{Dyer2016} and in-order parser by \citet{Liu2017}, where parsers are trained with a traditional static oracle. Both implementations follow a stack-LSTM approach to represent the stack and the buffer, as well as a vanilla LSTM to represent the action history. In addition, they also use a bi-LSTM
as a compositional function for representing constituents in the stack. Concretely, this consists in computing the composition representation $s_{comp}$ as:
\begin{multline*}
s_{comp} = (LSTM_{fwd}[e_{nt}, s_0, ... , s_m];\\
LSTM_{bwd}[e_{nt}, s_m, ... , s_0])
\end{multline*}
\noindent where $e_{nt}$ is the vector representation of a non-terminal, and $s_i$, $i \in [0,m]$ is the $i$th child node.

Finally, the exact same word representation strategy and hyper-parameter values as \cite{Dyer2016} and \cite{Liu2017} are used to conduct the experiments.

\subsection{Error exploration}
In order to benefit from training a parser by a dynamic oracle, errors should be made during the training process so that the parser can learn to avoid and recover from them. Unlike more complex error-exploration strategies as those studied in \cite{Ballesteros2016,Cross2016B,FriedK18}, we decided to consider a simple one that follows a non-optimal transition when it is the highest-scoring one, but with a certain probability. In that way, we easily simulate test time conditions, when the parser greedily chooses the highest-scoring transition, even when it is not an optimal one, placing the parser in an incorrect state.

In particular, we run experiments on development sets for each benchmark/algorithm with three different error exploration probabilities and choose the one that achieves the best F-score. Table~\ref{tab:error} reports all results, including those obtained by the top-down and in-order parsers trained by a dynamic oracle without error exploration (equivalent to a traditional static oracle).

 \begin{table}
 \begin{center}
 \begin{tabular}{@{\hskip 0.1pt}lcccc@{\hskip 0.1pt}}
  & \multicolumn{2}{c}{Top-down} & \multicolumn{2}{c}{In-order} \\
 Exp. & WSJ & CTB & WSJ & CTB \\
 \hline
  \small{None} & 91.81 & 88.94 & 91.95 & 89.69 \\
 \small{0.1} & 91.87 & 89.13 & \textbf{92.05} & \textbf{89.91} \\
 \small{0.2} & \textbf{91.99} & 88.70 & 91.98 & 89.88 \\
 \small{0.3} & 91.97 & \textbf{89.20} & 91.95 & 89.87 \\
 \hline
 \multicolumn{1}{c}{}\\
 \end{tabular}
 \caption{F-score comparison of different error-exploration probabilities on WSJ~\S22 and CTB~\S301-325 for the top-down a in-order dynamic oracles.}
 \label{tab:error}
 \end{center}
 \end{table}

\begin{table}
\begin{small}
\begin{center}
\centering
\begin{tabular}{@{\hskip 0.1pt}lccc@{\hskip 0.1pt}}
Parser & Type & Strat & F1 \\
\hline
\cite{Cross2016A} & gs & bu & 90.0  \\
\cite{Cross2016B} & gs & bu & 91.0  \\
\cite{Cross2016B} & gd & bu & 91.3  \\
\cite{Liu2017}  & gs & bu & 91.3  \\
\defcitealias{nonbinary}{Fern\'andez-G and G\'omez-R, 2018}\citepalias{nonbinary} & gs &  bu & 91.5\\ 
\defcitealias{nonbinary}{Fern\'andez-G and G\'omez-R, 2018}\citepalias{nonbinary}  & gd &  bu & 91.7\\
\cite{Dyer2016} & gs & td & 91.2    \\
\textbf{This work} & gd & td & \textbf{91.7}    \\
\cite{Liu2017}  & gs & in & 91.8   \\
\textbf{This work}  & gd & in & \textbf{92.0}   \\
\hline
\cite{Zhu13} & b & bu & 90.4  \\
\cite{Watanabe15} & b & bu & 90.7  \\
\cite{Liu2017B} & b & bu & 91.7    \\
\cite{FriedK18} & bp & td & 91.6    \\
\cite{FriedK18} & bd & td & 92.1    \\
\cite{FriedK18} & bp & in & 92.2    \\
\cite{SternFK17} & bg & td & 92.6 \\
\hline
\cite{SternAK17} & ch & bu & 91.8 \\
\cite{GaddySK18} & ch & bu & 92.1 \\
\cite{KleinK18} & ch & bu & \textbf{93.6} \\
\hline
\multicolumn{1}{c}{}\\
\end{tabular}

\begin{tabular}{@{\hskip 0pt}lccc@{\hskip 0pt}}

Parser & Type & Strat & F1 \\
\hline
\cite{Wang2015} & gs & bu & 83.2  \\
\cite{Liu2017}  & gs & bu & 85.7  \\
\defcitealias{nonbinary}{Fern\'andez-G and G\'omez-R, 2018}\citepalias{nonbinary} & gs &  bu & 86.3\\
\defcitealias{nonbinary}{Fern\'andez-G and G\'omez-R, 2018}\citepalias{nonbinary} & gd &  bu & 86.8\\

\cite{Dyer2016} & gs & td & 84.6    \\
\textbf{This work} & gd & td & \textbf{85.3}    \\
\cite{Liu2017}  & gs & in & 86.1   \\
\textbf{This work} & gd & in & \textbf{86.6}   \\
\hline
\cite{Zhu13} & b & bu & 83.2  \\
\cite{Watanabe15} & b & bu & 84.3  \\
\cite{Liu2017B} & b & bu & 85.5    \\
\cite{FriedK18} & bd & td & 85.5    \\
\cite{FriedK18} & bp & td & 84.7    \\
\cite{FriedK18} & bp & in & \textbf{87.0}    \\
\hline
\multicolumn{1}{c}{}\\
\end{tabular}

\caption{Accuracy comparison of state-of-the-art single-model fully-supervised constituent parsers on WSJ~\S{23} (top) and CTB~\S271-300 (bottom). The ``Type'' column shows the type of parser: \emph{gs} is a greedy parser trained with a static oracle, \emph{gd} a greedy parser trained with a dynamic oracle, \emph{b} a beam search parser, \emph{bp} a beam search parser trained with a policy gradient method, \emph{bd} a beam search parser trained with a non-optimal dynamic oracle, \emph{bg} a generative beam search parser, and \emph{ch} a chart-based parser. Finally, the ``Strat'' column describes the strategy followed ($bu$=bottom-up, $td$=top-down and $in$=in-order). 
}
\label{tab:results}

\end{center}
\end{small}
\end{table}

\begin{table*}[h]

\centering
\begin{tabular}{@{\hskip 0.1pt}lllllll@{\hskip 0.1pt}}
Parser & Oracle &\#1 & \#2 & \#3 & \#4 & \#5 \\
\hline
Top-down & static & 90.98 & 88.76 & 85.01 & 76.63 & 77.35 \\
 & dynamic & \textbf{91.34}{\scriptsize\ (+0.36)} & \textbf{89.18}{\scriptsize\ (+0.42)} & \textbf{85.17}{\scriptsize\ (+0.16)} & \textbf{77.12}{\scriptsize\ (+0.49)} & \textbf{80.02}{\scriptsize\ (+2.67)} \\
\hline
In-order & static & 91.36 & 89.21 & 85.15 & 77.08 & 79.02 \\
& dynamic & \textbf{91.55}{\scriptsize\ (+0.19)} & \textbf{89.43}{\scriptsize\ (+0.22)} &  \textbf{85.34}{\scriptsize\ (+0.19)} & \textbf{77.57}{\scriptsize\ (+0.49)} & \textbf{81.03}{\scriptsize\ (+2.01)} \\
\hline
\multicolumn{1}{c}{}\\
\end{tabular}
\caption{F-score on constituents with a number of children ranging from one to five on WSJ~\S23. 
}
\label{tab:analysis}
\end{table*}

\subsection{Results}
Table~\ref{tab:results} compares
our system's accuracy to other state-of-the-art shift-reduce constituent parsers on the WSJ and CTB benchmarks. For comparison, we also include some recent state-of-the-art parsers with global chart decoding that achieve the highest accuracies to date on WSJ, but are much slower than shift-reduce algorithms.

Top-down and in-order parsers benefit from being trained by these new dynamic oracles in both datasets. The top-down strategy achieves a gain of 0.5 and 0.7 points in F-score on WSJ and CTB benchmarks, respectively.
The in-order parser obtains similar improvements on the CTB (0.5 points), but less notable accuracy gain on the WSJ (0.2 points). Although a case of diminishing returns might explain the latter, 
the in-order parser trained with the proposed dynamic oracle still achieves the highest accuracy to date in greedy transition-based constituent parsing on the WSJ.\footnote{Note that the proposed dynamic oracles are orthogonal to 
approaches like beam search, re-ranking or semi-supervision, that can boost accuracy but at a large cost to parsing speed.}

While this work was under review,
\citet{FriedK18} proposed to train the top-down and in-order parsers with a policy gradient method instead of custom designed dynamic oracles. They also present a non-optimal dynamic oracle for 
the top-down parser that, combined with more complex error-exploration strategies and 
size-10 beam search, significantly outperforms the policy gradient-trained
version, confirming that even non-optimal dynamic oracles are 
a good
option.\footnote{Unfortunately, we cannot directly compare our approach to theirs, since they use 
beam-search decoding with size 10 in all experiments, gaining up to 0.3 points in F-score, while penalizing speed with respect to greedy decoding.
However, 
by extrapolating the results above,
we hypothesize that our optimal dynamic oracles (especially the one designed for the in-order algorithm) 
with their same training and beam-search decoding setup
might  
achieve the best scores to date in shift-reduce parsing.
}

\subsection{Analysis}
Dan  Bikel's  randomized  parsing  evaluation  comparator \cite{bikel04thesis} was used to perform significance tests on precision 
and  recall  metrics on WSJ~\S23 and CTB~\S271-300. 
The top-down parser trained with dynamic oracles achieves statistically significant improvements ($p<0.05$) in precision both on the WSJ and CTB benchmarks, and in recall on WSJ. The in-order parser trained with the proposed technique obtains significant improvements ($p<0.05$) in recall in both benchmarks, although not in precision.

We also undertake an analysis to check if dynamic oracles are able to mitigate error propagation.
We report in Table~\ref{tab:analysis} the F-score obtained in constituents with different number of children on WSJ~\S23 by the top-down and in-order algorithms trained with both static and dynamic oracles. Please note that creating a constituent with a great number of children is more prone to suffer from error propagation, since a larger number of transitions is required to build it. 
The results seem to confirm
that, indeed, dynamic oracles manage to alleviate error propagation, since improvements in F-score are more notable 
for larger constituents.

\section{Conclusion}
We develop the first optimal dynamic oracles for training the top-down and the state-of-the-art in-order parsers. Apart from improving the systems' accuracies in both cases, we achieve the best result to date in 
greedy shift-reduce parsing on the WSJ.
In addition, these promising techniques could easily benefit from recent studies in error-exploration strategies and yield  state-of-the-art accuracies in transition-based parsing in the near future.
The parser's source code is freely available at \url{https://github.com/danifg/Dynamic-InOrderParser}.

\section*{Acknowledgments}

This work has received funding from the European
Research Council (ERC), under the European
Union's Horizon 2020 research and innovation
programme (FASTPARSE, grant agreement No
714150), from 
MINECO (FFI2014-51978-C2-2-R, TIN2017-85160-C2-1-R)
and from Xunta de Galicia (ED431B 2017/01).

\bibliography{main,twoplanaracl,bibliography}

\newcommand{\beeksort}[1]{}
\begin{thebibliography}{26}
\expandafter\ifx\csname natexlab\endcsname\relax\def\natexlab#1{#1}\fi

\bibitem[{Ballesteros et~al.(2016)Ballesteros, Goldberg, Dyer, and
  Smith}]{Ballesteros2016}
Miguel Ballesteros, Yoav Goldberg, Chris Dyer, and Noah~A. Smith. 2016.
\newblock \href {http://aclweb.org/anthology/D/D16/D16-1211.pdf} {Training with
  exploration improves a greedy stack {LSTM} parser}.
\newblock In \emph{Proceedings of the 2016 Conference on Empirical Methods in
  Natural Language Processing, {EMNLP} 2016, Austin, Texas, USA, November 1-4,
  2016}, pages 2005--2010.

\bibitem[{Bikel(2004)}]{bikel04thesis}
Dan Bikel. 2004.
\newblock \emph{On the Parameter Space of Generative Lexicalized Statistical
  Parsing Models}.
\newblock Ph.D. thesis, University of Pennsylvania.

\bibitem[{Coavoux and Crabb\'{e}(2016)}]{Coavoux2016}
Maximin Coavoux and Benoit Crabb\'{e}. 2016.
\newblock \href {http://www.aclweb.org/anthology/P16-1017} {Neural greedy
  constituent parsing with dynamic oracles}.
\newblock In \emph{Proceedings of the 54th Annual Meeting of the Association
  for Computational Linguistics (Volume 1: Long Papers)}, pages 172--182,
  Berlin, Germany. Association for Computational Linguistics.

\bibitem[{Crabb\'{e}(2015)}]{Crabbe2015}
Benoit Crabb\'{e}. 2015.
\newblock \href {http://aclweb.org/anthology/D15-1212} {Multilingual
  discriminative lexicalized phrase structure parsing}.
\newblock In \emph{Proceedings of the 2015 Conference on Empirical Methods in
  Natural Language Processing}, pages 1847--1856, Lisbon, Portugal. Association
  for Computational Linguistics.

\bibitem[{Cross and Huang(2016{\natexlab{a}})}]{Cross2016A}
James Cross and Liang Huang. 2016{\natexlab{a}}.
\newblock Incremental parsing with minimal features using bi-directional
  {LSTM}.
\newblock In \emph{{ACL} {(2)}}. The Association for Computer Linguistics.

\bibitem[{Cross and Huang(2016{\natexlab{b}})}]{Cross2016B}
James Cross and Liang Huang. 2016{\natexlab{b}}.
\newblock Span-based constituency parsing with a structure-label system and
  provably optimal dynamic oracles.
\newblock In \emph{{EMNLP}}, pages 1--11. The Association for Computational
  Linguistics.

\bibitem[{Dyer et~al.(2016)Dyer, Kuncoro, Ballesteros, and Smith}]{Dyer2016}
Chris Dyer, Adhiguna Kuncoro, Miguel Ballesteros, and Noah~A. Smith. 2016.
\newblock Recurrent neural network grammars.
\newblock In \emph{{HLT-NAACL}}, pages 199--209. The Association for
  Computational Linguistics.

\bibitem[{Fern\'andez-Gonz\'alez and G\'omez-Rodr\'iguez(2018)}]{nonbinary}
Daniel Fern\'andez-Gonz\'alez and Carlos G\'omez-Rodr\'iguez. 2018.
\newblock \href {http://arxiv.org/abs/arXiv:1804.07961} {Faster shift-reduce
  constituent parsing with a non-binary, bottom-up strategy}.
\newblock \emph{arXiv}, 1804.07961 [cs.CL].

\bibitem[{Fredman(1975)}]{Fredman1975}
Michael~L. Fredman. 1975.
\newblock \href {https://doi.org/https://doi.org/10.1016/0012-365X(75)90103-X}
  {On computing the length of longest increasing subsequences}.
\newblock \emph{Discrete Mathematics}, 11(1):29 -- 35.

\bibitem[{Fried and Klein(2018)}]{FriedK18}
Daniel Fried and Dan Klein. 2018.
\newblock \href {https://aclanthology.info/papers/P18-2075/p18-2075} {Policy
  gradient as a proxy for dynamic oracles in constituency parsing}.
\newblock In \emph{Proceedings of the 56th Annual Meeting of the Association
  for Computational Linguistics, {ACL} 2018, Melbourne, Australia, July 15-20,
  2018, Volume 2: Short Papers}, pages 469--476.

\bibitem[{Gaddy et~al.(2018)Gaddy, Stern, and Klein}]{GaddySK18}
David Gaddy, Mitchell Stern, and Dan Klein. 2018.
\newblock \href {https://aclanthology.info/papers/N18-1091/n18-1091} {What's
  going on in neural constituency parsers? an analysis}.
\newblock In \emph{Proceedings of the 2018 Conference of the North American
  Chapter of the Association for Computational Linguistics: Human Language
  Technologies, {NAACL-HLT} 2018, New Orleans, Louisiana, USA, June 1-6, 2018,
  Volume 1 (Long Papers)}, pages 999--1010.

\bibitem[{Goldberg and Nivre(2012)}]{goldberg2012dynamic}
Yoav Goldberg and Joakim Nivre. 2012.
\newblock \href {http://www.aclweb.org/anthology/C12-1059} {A dynamic oracle
  for arc-eager dependency parsing}.
\newblock In \emph{Proceedings of COLING 2012}, pages 959--976, Mumbai, India.
  Association for Computational Linguistics.

\bibitem[{Kitaev and Klein(2018)}]{KleinK18}
Nikita Kitaev and Dan Klein. 2018.
\newblock \href {https://aclanthology.info/papers/P18-1249/p18-1249}
  {Constituency parsing with a self-attentive encoder}.
\newblock In \emph{Proceedings of the 56th Annual Meeting of the Association
  for Computational Linguistics, {ACL} 2018, Melbourne, Australia, July 15-20,
  2018, Volume 1: Long Papers}, pages 2675--2685.

\bibitem[{Kuncoro et~al.(2017)Kuncoro, Ballesteros, Kong, Dyer, Neubig, and
  Smith}]{Kuncoro2017}
Adhiguna Kuncoro, Miguel Ballesteros, Lingpeng Kong, Chris Dyer, Graham Neubig,
  and Noah~A. Smith. 2017.
\newblock What do recurrent neural network grammars learn about syntax?
\newblock In \emph{{EACL} {(1)}}, pages 1249--1258. Association for
  Computational Linguistics.

\bibitem[{Liu and Zhang(2017{\natexlab{a}})}]{Liu2017}
Jiangming Liu and Yue Zhang. 2017{\natexlab{a}}.
\newblock \href {https://www.transacl.org/ojs/index.php/tacl/article/view/1199}
  {In-order transition-based constituent parsing}.
\newblock \emph{Transactions of the Association for Computational Linguistics},
  5:413--424.

\bibitem[{Liu and Zhang(2017{\natexlab{b}})}]{Liu2017B}
Jiangming Liu and Yue Zhang. 2017{\natexlab{b}}.
\newblock Shift-reduce constituent parsing with neural lookahead features.
\newblock \emph{{TACL}}, 5:45--58.

\bibitem[{Marcus et~al.(1993)Marcus, Santorini, and Marcinkiewicz}]{marcus93}
Mitchell~P. Marcus, Beatrice Santorini, and Mary~Ann Marcinkiewicz. 1993.
\newblock Building a large annotated corpus of {E}nglish: The {P}enn
  {T}reebank.
\newblock \emph{Computational Linguistics}, 19:313--330.

\bibitem[{Mi and Huang(2015)}]{Mi2015}
Haitao Mi and Liang Huang. 2015.
\newblock \href {http://www.aclweb.org/anthology/N15-1108} {Shift-reduce
  constituency parsing with dynamic programming and pos tag lattice}.
\newblock In \emph{Proceedings of the 2015 Conference of the North American
  Chapter of the Association for Computational Linguistics: Human Language
  Technologies}, pages 1030--1035, Denver, Colorado. Association for
  Computational Linguistics.

\bibitem[{Sagae and Lavie(2005)}]{sagae05}
Kenji Sagae and Alon Lavie. 2005.
\newblock A classifier-based parser with linear run-time complexity.
\newblock In \emph{Proceedings of the 9th International Workshop on Parsing
  Technologies (IWPT)}, pages 125--132.

\bibitem[{Stern et~al.(2017{\natexlab{a}})Stern, Andreas, and
  Klein}]{SternAK17}
Mitchell Stern, Jacob Andreas, and Dan Klein. 2017{\natexlab{a}}.
\newblock \href {https://doi.org/10.18653/v1/P17-1076} {A minimal span-based
  neural constituency parser}.
\newblock In \emph{Proceedings of the 55th Annual Meeting of the Association
  for Computational Linguistics, {ACL} 2017, Vancouver, Canada, July 30 -
  August 4, Volume 1: Long Papers}, pages 818--827.

\bibitem[{Stern et~al.(2017{\natexlab{b}})Stern, Fried, and Klein}]{SternFK17}
Mitchell Stern, Daniel Fried, and Dan Klein. 2017{\natexlab{b}}.
\newblock \href {https://aclanthology.info/papers/D17-1178/d17-1178} {Effective
  inference for generative neural parsing}.
\newblock In \emph{Proceedings of the 2017 Conference on Empirical Methods in
  Natural Language Processing, {EMNLP} 2017, Copenhagen, Denmark, September
  9-11, 2017}, pages 1695--1700.

\bibitem[{Wang et~al.(2015)Wang, Mi, and Xue}]{Wang2015}
Zhiguo Wang, Haitao Mi, and Nianwen Xue. 2015.
\newblock Feature optimization for constituent parsing via neural networks.
\newblock In \emph{Proceedings of the 53rd Annual Meeting of the Association
  for Computational Linguistics and the 7th International Joint Conference on
  Natural Language Processing of the Asian Federation of Natural Language
  Processing, {ACL} 2015, July 26-31, 2015, Beijing, China, Volume 1: Long
  Papers}, pages 1138--1147.

\bibitem[{Watanabe and Sumita(2015)}]{Watanabe15}
Taro Watanabe and Eiichiro Sumita. 2015.
\newblock Transition-based neural constituent parsing.
\newblock In \emph{Proceedings of the 53rd Annual Meeting of the Association
  for Computational Linguistics, {ACL} 2015, 26-31 July 2015, Bejing, China,
  Volume 1: Long Papers}, pages 1169--1179.

\bibitem[{Xue et~al.(2005)Xue, Xia, Chiou, and Palmer}]{Xue2005}
Naiwen Xue, Fei Xia, Fu-dong Chiou, and Marta Palmer. 2005.
\newblock \href {https://doi.org/10.1017/S135132490400364X} {The penn chinese
  treebank: Phrase structure annotation of a large corpus}.
\newblock \emph{Nat. Lang. Eng.}, 11(2):207--238.

\bibitem[{Zhang and Clark(2009)}]{Zhang2009}
Yue Zhang and Stephen Clark. 2009.
\newblock \href {http://dl.acm.org/citation.cfm?id=1697236.1697267}
  {Transition-based parsing of the chinese treebank using a global
  discriminative model}.
\newblock In \emph{Proceedings of the 11th International Conference on Parsing
  Technologies}, IWPT '09, pages 162--171, Stroudsburg, PA, USA. Association
  for Computational Linguistics.

\bibitem[{Zhu et~al.(2013)Zhu, Zhang, Chen, Zhang, and Zhu}]{Zhu13}
Muhua Zhu, Yue Zhang, Wenliang Chen, Min Zhang, and Jingbo Zhu. 2013.
\newblock \href {http://aclweb.org/anthology/P/P13/P13-1043.pdf} {Fast and
  accurate shift-reduce constituent parsing}.
\newblock In \emph{Proceedings of the 51st Annual Meeting of the Association
  for Computational Linguistics, {ACL} 2013, 4-9 August 2013, Sofia, Bulgaria,
  Volume 1: Long Papers}, pages 434--443.

\end{thebibliography}
\bibliographystyle{acl_natbib}

\end{document}